# High-speed real-time single-pixel microscopy based on Fourier sampling

QIANG GUO,[1] HONGWEI CHEN,[1,*] YUXI WANG,[1] YONG GUO,[2] PENG LIU,[2] XIURUI ZHU,[2] ZHENG CHENG,[2] ZHENMING YU,[1] MINGHUA CHEN,[1] SIGANG YANG,[1] AND SHIZHONG XIE[1]

[1]*Tsinghua National Laboratory for Information Science and Technology (TNList), Department of Electronic Engineering, Tsinghua University, Beijing, 100084, China*
[2]*Department of Biomedical Engineering, School of Medicine, Collaborative Innovation Center for Diagnosis and Treatment of Infectious Diseases, Tsinghua University, Beijing, 100084, China*
*\*Corresponding author: chenhw@mail.tsinghua.edu.cn*

**Single-pixel cameras based on the concepts of compressed sensing (CS) leverage the inherent structure of images to retrieve them with far fewer measurements and operate efficiently over a significantly broader spectral range than conventional silicon-based cameras. Recently, photonic time-stretch (PTS) technique facilitates the emergence of high-speed single-pixel cameras. A significant breakthrough in imaging speed of single-pixel cameras enables observation of fast dynamic phenomena. However, according to CS theory, image reconstruction is an iterative process that consumes enormous amounts of computational time and cannot be performed in real time. To address this challenge, we propose a novel single-pixel imaging technique that can produce high-quality images through rapid acquisition of their effective spatial Fourier spectrum. We employ phase-shifting sinusoidal structured illumination instead of random illumination for spectrum acquisition and apply inverse Fourier transform to the obtained spectrum for image restoration. We evaluate the performance of our prototype system by recognizing quick response (QR) codes and flow cytometric screening of cells. A frame rate of 625 kHz and a compression ratio of 10% are experimentally demonstrated in accordance with the recognition rate of the QR code. An imaging flow cytometer enabling high-content screening with an unprecedented throughput of 100,000 cells/s is also demonstrated. For real-time imaging applications, the proposed single-pixel microscope can significantly reduce the time required for image reconstruction by two orders of magnitude, which can be widely applied in industrial quality control and label-free biomedical imaging. © 2016 Optical Society of America**

http://dx.doi.org/10.1364/optica.99.099999

## 1. INTRODUCTION

Computational imaging (CI) based on optical encoding is a promising technique for image acquisition with spatially unresolved detectors and can offer a significantly broader spectral range than conventional silicon-based cameras using a charge-coupled device (CCD) or complementary metal-oxide semiconductor (CMOS) [1]. Ghost imaging (GI), as an original technique, uses correlation measurements between the light transmitted through (or reflected by) an object and the spatially resolved intensity pattern of the incident light to reconstruct the ghost image of the object [2-15]. However, large amounts of measurements are required to produce high-quality images, which seriously limits the imaging speed. By exploiting the sparsity property of images, single-pixel imaging (SPI) based on the theory of compressed sensing (CS) enables image reconstruction with far fewer measurements than is possible with conventional GI [16-20]. In a typical single-pixel imaging system, a digital micromirror device (DMD) is employed to perform optical random encryption and its maximum refresh rate of 22 kHz puts a strict limit on the frame rate, which is a barrier for observing fast dynamic phenomena. In recent years, photonic time-stretch technique (PTS) is applied to achieve ultrafast real-time optical imaging for capturing fast dynamic events, making it possible to overcome the compromise between sensitivity and frame rate [21-24]. An imaging flow cytometer with an unprecedented throughput of 100,000 particles/s for circulating tumor cells (CTCs) detection is also demonstrated [23]. However, dozens of gigabytes of data per second produced by the imaging flow cytometer is a critical problem that overwhelms its data acquisition and processing backend. To solve this challenging problem and simultaneously improve the imaging speed of single-pixel cameras, the first high-speed single-pixel imaging system based on photonic time stretch is implemented by integrating time-stretch technique and CS theory [25-28]. High-speed electro-optic modulators replacing DMDs are used to perform optical random encoding, resulting in an increase in imaging speed of approximately three orders of magnitude. Currently, single-pixel imaging systems with multi-megahertz frame rates have been demonstrated, enabling capture of microscopic objects moving at high speed [27]. In all these systems, compressive measurements are acquired at a rate amounting to the repetition rate of the optical pulses, which can greatly relieve pressure on sampling and storage. However, image reconstruction is an iterative optimization process and it suffers from extremely time-consuming computations, which is a major bottleneck for a real-time imaging implementation.

Here, we exploit a unique property that the Fourier representations of natural images are sparse and most of their energy is concentrated in the low spatial frequency region. Extracting the effective frequency information and discarding the high-frequency components allows us to perform data compression without introducing a large distortion to the image. Meanwhile, images can be reconstructed by applying inverse discrete Fourier transform (IDFT) to the captured measurements, which is an existing module in current digital signal processors (DSPs). According to the above analysis, we design a high-speed single-pixel imaging system that can produce high-quality images by only acquiring their effective Fourier spectrum. Four-step phase-shifting sinusoidal light projection is employed for spectrum acquisition [29, 30]. We first use our prototype system to recognize quick response (QR) codes for performance evaluation. A frame rate of 625 kHz and a compression ratio of 10% are experimentally demonstrated in accordance with the recognition rate of the QR codes. Besides, an imaging cytometer with a throughput of 100,000 cells/s for high-content screening is implemented. Compared with previous techniques, our scheme can significantly reduce the computational time required for image reconstruction, which contributes to achieving real-time imaging. In the following, we will describe the basic principle of our approach, present the system prototype and demonstrate real-time compressive microscopy by imaging various small moving objects.

## 2. METHODS

Any image can be decomposed into a sum of harmonic sinusoidal patterns with different spatial frequencies and amplitudes. If the weight of each spatial frequency is known, the image can be reconstructed faithfully. It is a remarkable fact that the energy of a natural image is concentrated in the low spatial frequency region and consequently it is possible to perform image reconstruction and compression simultaneously by only extracting the low spatial frequency components of the image as shown in Fig. 1(a). The previous CS-based approach can perform image reconstruction without knowing the exact locations of the sparse components in the spatial frequency domain in advance as shown in Fig. 1(b). However, the reconstruction process requires a large number of iterations and is quite time-consuming. Compared with the CS-based scheme, our scheme utilizes the prior information that the sparse frequency components of natural images mainly locate at low spatial frequencies to capture the effective spectrum and then performs an IDFT operation on the obtained spectrum for image reconstruction. As the IDFT operation is quite computationally efficient, this scheme is a promising approach to achieving real-time image reconstruction. The procedure for computing the Fourier transform is shown as Eq. (1).

$$\begin{bmatrix} X(0) \\ X(1) \\ X(2) \\ \vdots \\ X(N-1) \end{bmatrix} = \begin{bmatrix} 1 & 1 & 1 & \cdots & 1 \\ 1 & W_N^1 & W_N^2 & \cdots & W_N^{(N-1)} \\ 1 & W_N^2 & W_N^4 & \cdots & W_N^{2(N-1)} \\ \vdots & \vdots & \vdots & \ddots & \vdots \\ 1 & W_N^{N-1} & W_N^{2(N-1)} & \cdots & W_N^{(N-1)(N-1)} \end{bmatrix} \begin{bmatrix} x(0) \\ x(1) \\ x(2) \\ \vdots \\ x(N-1) \end{bmatrix} \quad (1)$$

where $x(n)$ ($n=0,1,\cdots N-1$) denotes a signal and $X(k)$ ($k=0,1,\cdots N-1$) is its Fourier transformation. $W_N^K = e^{-j\frac{2\pi}{N}K} = \cos(\frac{2\pi}{N}K) - j\sin(\frac{2\pi}{N}K)$ (K=0,1,···N-1). As can be seen in Eq. (1), the weight of each frequency is a complex number and its real part (or imaginary part) is the inner product of the signal and a cosinoidal signal (or a sinusoidal signal) at the corresponding frequency. Motivated by this conception, we can obtain the spatial spectrum of a natural image with a phase-shifting method as shown in Fig. 1(c) and then reconstruct the image by performing IDFT.

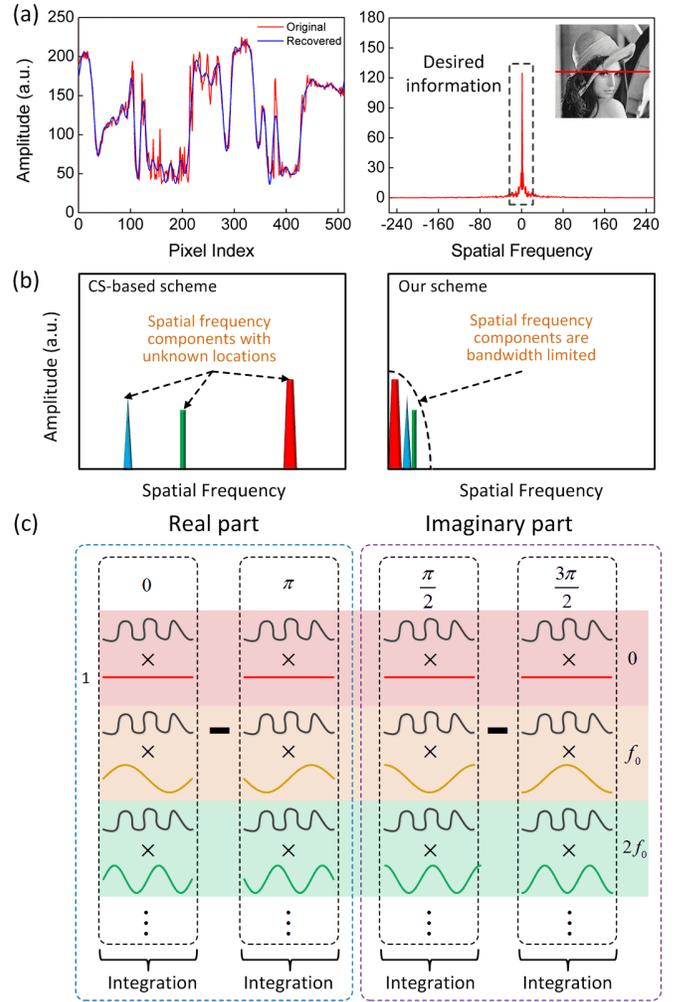

**Fig. 1.** (a) The sparse property of natural images and the corresponding image compression scheme. (b) Comparison of the CS-based scheme and our scheme. (c) The detailed process of the Fourier transformation. The curve (solid black line) represents a one-dimensional signal and the curves (solid red line, yellow line and green line) are the sinusoidal signals with different frequencies.

To achieve discrete Fourier transform (DFT) (shown in Fig. 1(c)) in the optical domain, a mode locked laser (MLL) with a repetition rate of $f_{rep}$ is used as the light source. Through a temporal dispersive element, the spectrum of each pulse is mapped to a temporal waveform. Then we can perform optical spectral shaping with sinusoidal light patterns whose frequencies are harmonics of the fundamental frequency $f_0$ ($f_0 = f_{rep}$). The applied four-step phase-shifting patterns are specified with the same frequency $f_i = i \cdot f_0$ and four initial phases $\varphi$ (0, $\pi/2$, $\pi$ and $3\pi/2$), respectively. Through optical intensity modulation, the output light intensity is given by:

$$P_{out}(t, f_i) = \frac{1}{2} P_{in}(t)[1 + \alpha \cos(2\pi f_i t + \varphi)] \quad (2)$$

where $P_{in}(t)$ denotes the input optical pulse train. $\alpha = \pi / V_\pi$ is the modulation index of the electro-optic intensity modulator, where $V_\pi$ is the half-wave voltage. The time-to-frequency conversion (also known as dispersive Fourier transformation) is mainly a linear mapping [22], and consequently the intensity of the dispersed pulse mimics its spectrum. Then a diffraction grating is used to produce the corresponding spatial patterns to illuminate the scene. As the frequency-to-space mapping is also a linear process [22], we can

produce one-dimensional (1-D) sinusoidal light patterns which can be written as:

$$C_{out}(x, f_x) = A + B\cos(2\pi f_x x + \varphi) \quad (3)$$

where $A$ and $B$ account for the average brightness and the fringe contrast, respectively. $f_x$ is the corresponding spatial frequency resulting from time-to-frequency-to-space mapping, where $x$ denotes the 1-D Cartesian coordinate in the scene. After illuminating the scene, pulse compression is performed and the measurements (namely the pulse energy) are captured by a bucket detector. Herein, the compressive measurements are expressed as:

$$V_\varphi(f_x) = K\int_x I(x) C_{out}(x, f_x) dx + V_n \quad (4)$$

where $I(x)$ is the intensity distribution of the illuminated scene. $K$ is a constant and $V_n$ is the noise term. Each complex Fourier coefficient can be obtained from four measurements ($V_0$, $V_{\pi/2}$, $V_\pi$ and $V_{3\pi/2}$) with DC offset cancelation as follows:

$$[V_0(f_x) - V_\pi(f_x)] + j \cdot [V_{\pi/2}(f_x) - V_{3\pi/2}(f_x)] = K\omega_{f_x} \quad (5)$$

where $\omega_{f_x}$ is the weight of the spatial frequency $f_x$. In this way, we can obtain the spatial spectrum of the 1-D image $I(x)$. Finally IDFT is performed to reconstruct the original 1-D image.

$$I(x) = \frac{1}{K} F^{-1}\{[\omega_0, \omega_{f_0}, \omega_{2f_0}, \cdots]\} \quad (6)$$

where $F^{-1}$ denotes the inverse Fourier transform operator and $[\omega_0, \omega_{f_0}, \omega_{2f_0}, \cdots]$ denotes a vector composed of the weights of different spatial frequencies.

## 3. SYSTEM CONFIGURATION

### A. Experimental architecture

The schematic diagram of the proposed high-speed single-pixel imaging system based on Fourier spectrum acquisition is shown in Fig. 1. A passively mode-locked laser (MLL) (PriTel HPPRR-FFL-50MHz) operating at 1550 nm is used as the light source that generates 150-fs optical pulses with an average power of 32 mW and a spectral width of 15 nm at a repetition rate of 50 MHz. The optical pulses pass through a section of dispersion compensating fiber (DCF) with a group velocity dispersion (GVD) of -1368 ps/nm. The spectrum of each pulse is mapped to a temporal waveform whose duration is approximately 20 ns. To compensate optical loss induced by the DCF, a high-power erbium-doped fiber amplifier (EDFA) (Amonics AEDFA-33-B-FA) with a saturated output power of 33 dBm is employed to amplify the dispersed pulses. Then we use an arbitrary waveform generator (AWG) (Tektronix AWG70000A) which is synchronized to the MLL to generate phase-shifting sinusoidal patterns for spectral encoding of the pulses through a 12.5-Gb/s Mach-Zehnder modulator (MZM) (PHOTLINE MX-LN-10). The fundamental frequency of the sinusoidal patterns is equal to the pulse repetition rate (50 MHz). Sinusoidal patterns of the same frequency with four different phases (0, $\pi/2$, $\pi$ and $3\pi/2$) are used to sequentially encode the spectrum of four adjacent pulses. After sinusoidal modulation, an optical circulator directs the spectrally encoded pulses into a diffraction grating with a groove density of 1,200 lines per millimeter. Due to the angular dispersion of the diffraction grating, a 1-D rainbow pattern (50 µm long) for line-scan imaging is produced via wavelength-to-space mapping as shown in the inset of Fig. 2. The detailed structures of the scan lines with different spatial frequencies ($f$ and $2f$) and four phases (0, $\pi/2$, $\pi$ and $3\pi/2$) are depicted in Fig. 3, respectively. The figures on the left show the intensity curves versus the position on the scan line and the figures on the right are the corresponding spatial patterns of the scan lines, exhibiting sinusoidal intensity distributions. Passing through an objective lens with a focal length of 10 mm (Mitutoyo, 20×, M Plan Apo NIR, NA 0.40), the resultant 1-D rainbow beam is focused on the samples (a designed QR code and various flowing cells are tested). Then the spectrum of each pulse is encoded by the spatial information of the sample. The reflected pulses from the sample re-enter the diffraction grating followed by the circulator. A section of single mode fiber (SMF) with a length of 80 km and a complementary GVD of 17 ps/nm/km is used to perform pulse compression. Finally the energy of the compressed pulses is detected by a 1.2-GHz photodiode (THORLABS DET01CFC/M) and sampled by an externally-clocked analog-to-digital converter (ADC) synchronized with a 50-MHz sampling clock derived from the MLL. The captured pulse energy is related to the weight of each spatial frequency of the image. Image reconstruction can be performed in real time by only applying IDFT to the acquired measurements.

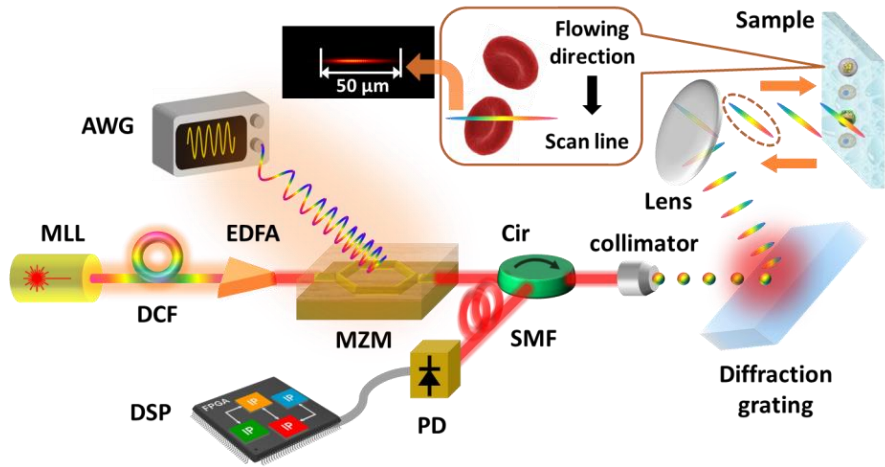

**Fig. 2.** Experimental setup. MLL, mode-locked laser; DCF, dispersion compensating fiber; EDFA, erbium-doped fiber amplifier; MZM, Mach-Zehnder modulator; AWG, arbitrary waveform generator; Cir, circulator; SMF, single mode fiber; PD, photo-detector; DSP, digital signal processor. The inset: the generated 1-D rainbow pattern with a length of approximately 50 µm.

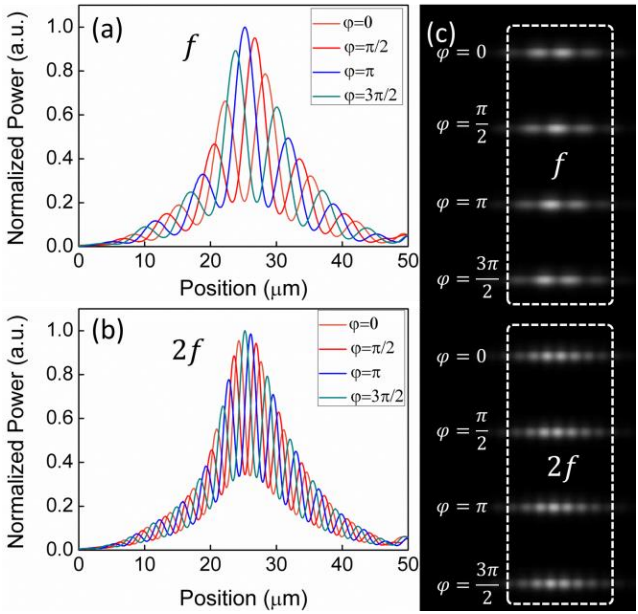

**Fig. 3.** The detailed structures of the sinusoidal light patterns with different spatial frequencies ($f$ and $2f$) and four phases ($0$, $\pi/2$, $\pi$ and $3\pi/2$).

### B. Experimental samples

In the experiment, a calibration target with a QR code pattern (21×21 pixels, 5 μm×5 μm pixel size) is used to evaluate the performance of our prototype system. The calibration target is placed normal to the incident beam on a motorized X–Y stage which moves the target to perform a scan across the whole QR code pattern. The step size of the motorized stage and the number of steps are set to 1 μm and 200, respectively.

To show the utility of our system, an imaging flow cytometer enabling high-throughput screening is also demonstrated. In the experiment, a microfluidic device is a necessity where the cells are controlled to flow at a uniform velocity, focused and ordered in the microfluidic channel by intrinsic inertial lift forces. Herein, the device is fabricated in thermoset polyester to ensure the rigidity of the microfluidic channel. It is important because the undesirable fluid dynamics and optical effects will be caused if the channel deforms. The dimensions of the microfluidic device used in the experiment are 120 μm width, 60 μm height and 2.2 cm length. The cells in the channel are controlled to flow at a speed of 1 m/s, which corresponds to a throughput of 100,000 cells/s.

### C. Experimental performance

There are two important indexes that are used to evaluate the performance of our prototype system: the frame rate $F$ and the compression ratio $R$.

Our system can enable image compression due to the sparsity property of the spatial Fourier spectrum of natural images. It is worth noting that we should maintain high-quality reconstructions when performing image compression. The compression ratio $R$ describes the capability to compress the data volume and is expressed as:

$$R = \frac{M}{N} \qquad (7)$$

where $M$ is the number of measurements used for image reconstruction. $N$ is defined as the number of pixels that a 1-D image contains. In the experiment, a digital storage oscilloscope (DSO) (Agilent Infiniium DSO91204A) with a sampling rate of 40 GS/s and a bandwidth of 13 GHz is used to digitize the temporal waveforms that carry the spatial information of the sample before pulse compression. Herein, $N = \Delta\lambda \cdot |D| \cdot f_s$, where $\Delta\lambda$ is the spectral width of the pulse (15 nm), $D$ denotes the dispersion of the DCF (-1368 ps/nm) and $f_s$ is the sampling rate of the DSO (40 GS/s). By substituting the corresponding parameters in Eq. (6), we can infer that $N$ is equal to 800.

On the other hand, the frame rate of our prototype system is referred to:

$$F = \frac{f_{rep}}{M} \qquad (8)$$

where $f_{rep}$ is the pulse repetition rate (50 MHz). From Eq. (8), it can be seen that the frame rate of our system $F$ is inversely proportional to the number of measurements $M$, whereas the compression ratio $R$ is proportional to the number of measurements $M$. Consequently, a trade-off between the frame rate and the compression ratio should be taken into consideration.

## 4. RESULTS

### A. QR code recognition

Surface defect and morphology inspection is one of the significantly applications in industry [31]. To verify the fact that our system has the capability of performing fast surface inspection in real time, we use our system to capture a photograph of a QR code (21×21 pixels, 5 μm×5 μm pixel size) and process the image with a QR code reader. The original image directly captured by the DSO is shown in Fig. 4(a) and the information encoded in the QR code is a line of text "THU-EE". Figs. 4(b)-4(f) give the reconstructed images at different compression ratios (25%, 15%, 10%, 7.5% and 5%). By using a QR code reader to identify the codes, the reconstructed QR code images at compression ratios over 10% can be well recognized. To quality the distortion of the reconstructed images as a function of the compression ratio, we calculate the mean squared error (MSE) and the peak signal-to-noise ratio (PSNR), respectively.

$$MSE = \frac{1}{mn}\sum_{i=1}^{m}\sum_{j=1}^{n}\|I(i,j) - K(i,j)\|^2 \qquad (9)$$

$$PSNR = 10 \cdot \lg(\frac{MAX_I^2}{MSE}) = 20\lg(\frac{MAX_I}{\sqrt{MSE}}) \qquad (10)$$

where $I(i,j)$ denotes the original image of size $m \times n$ captured by the DSO and $K(i,j)$ is the corresponding reconstructed image at a certain compression ratio. $MAX_I$ denotes the maximum possible pixel value of the image. As shown in Fig. 5, the calculated PSNR (23.09 dB, 22.71 dB, 21.51 dB, 18.23 and 16.66 dB) is positively correlated with the compression ratio (25%, 15%, 10%, 7.5% and 5%), which implies that better reconstructions will be achieved if more measurements are acquired for image restoration. The maximum movement velocity of the motorized stage is determined by the pixel size of the QR code image ($\Delta L$), the period of the optical pulses ($T$) and the minimum number of measurements used for image reconstruction ($M$).

$$v = \frac{\Delta L}{M \cdot T} \qquad (11)$$

Here $\Delta L$ is 5 μm, $T$ is 20 ns and 80 measurements are used. Consequently, the maximum movement velocity of the motorized stage is approximately 3.1 m/s. If we increase the pulse repetition rate, the moving QR code image with a larger velocity can be well reconstructed.

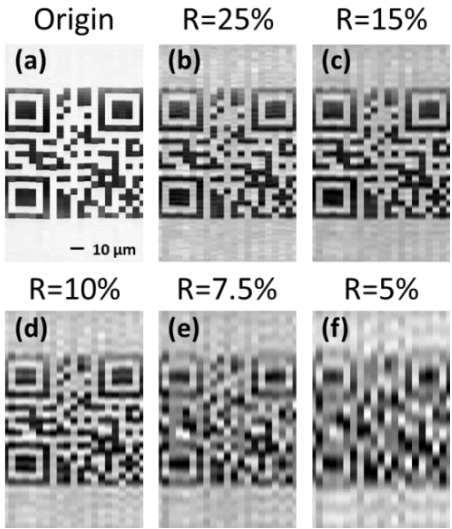

**Fig. 4.** The original QR code image (21×21 pixels) and reconstructed images at compression ratios equal to 25%, 15%, 10%, 7.5% and 5%.

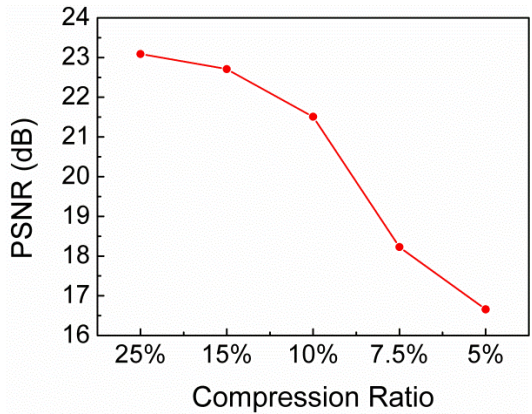

**Fig. 5.** Plot of PSNR versus compression ratio for QR code recognition.

To further evaluate recovery performance, a certain 1-D image extracted from the original two-dimensional (2-D) QR code image is compared with those from the reconstructed images at different compression ratios. Fig. 6 gives the original 1-D image (green curve), the corresponding reconstructed image at a compression ratio of 10% (red curve) and a compression ratio of 5% (blue curve). The results show that when the compression ratio is 5%, some useful information of the 1-D image has been lost, which leads to degradation of image quality. While at a compression ratio of 10%, the details of the image appear to be well recovered.

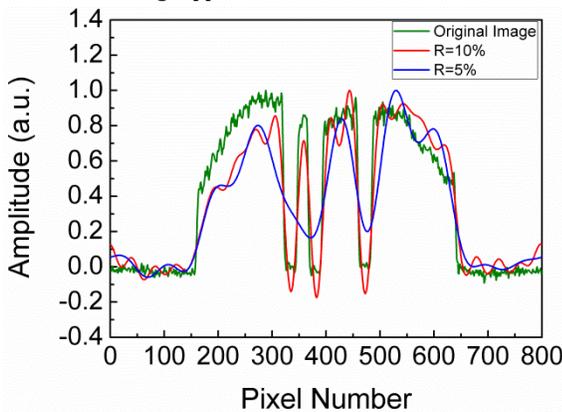

**Fig. 6.** Comparison between the original 1-D image and the reconstructed ones at compression ratios equal to 10% and 5%.

### B. Imaging flow cytometry

In order to investigate the potential of our prototype system for biological applications [32-34], we experimentally demonstrate an imaging flow cytometer with a throughput of 100,000 cells/s for high-throughput screening. In the experiment, various biological samples including oil droplets, red cells and Jurkat cells are imaged by our proposed imaging system. Fig. 7 shows the original and reconstructed snapshots of the flowing particles in the microfluidic channel. Here the particles are controlled to flow at a uniform speed of 1 m/s, which corresponds to a throughput of 100,000 cells/s. The figures on the left present the original images captured by the DSO without applying optical encoding and the figures on the right are the corresponding reconstructed images at a compression ratio of 10% (80 measurements). The PSNR of the reconstructed images are 21.28 dB, 21.22 dB and 21.32 dB, respectively, implying a reliable reconstruction of flowing cells. Moreover, the frame rate of the imaging flow cytometer is 625 kHz, which is sufficiently high for screening of flowing particles.

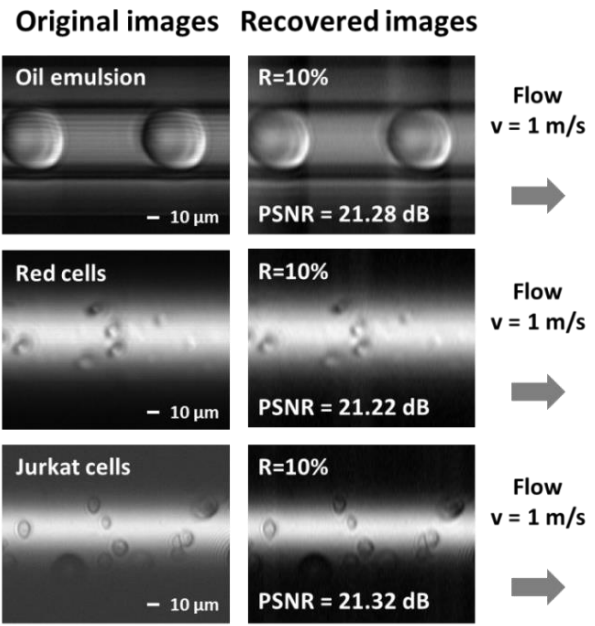

**Fig. 7.** Original (on the left) and reconstructed (on the right) E-slides of particles flowing at 1 m/s in the microfluidic channel. The compression ratio is 10% and the PSNR of the corresponding reconstructed images are also given.

Herein, we analyze the effect of the imaging frame rate and the flow speed of the particles on image reconstruction accuracy. We can assume that the image changes slowly across a group of scanlines with respect to the high repetition rate of femtosecond laser pulses (50 MHz). The number of scanlines assigned to a frame is determined by the imaging frame rate and the flow speed of the particles. As shown in Fig. 8, when 40 measurements (too small) are used to reconstruct a frame, some useful information of the 1-D image has been lost, leading to image quality degradation. Whereas when the number of measurements is too large (M=400), it cannot be ensured that the image will still remain nearly unchanged over a period of 8 μs (400 scan periods), which also degrades the quality of the reconstructed image. The results in Fig. 8 indicate that if the number of the acquired measurements is selected too small or too large, the recovery accuracy will be badly affected. Therefore, the number of measurements used for image reconstruction should be designed carefully, for the purpose of achieving accurate reconstruction and high frame rates.

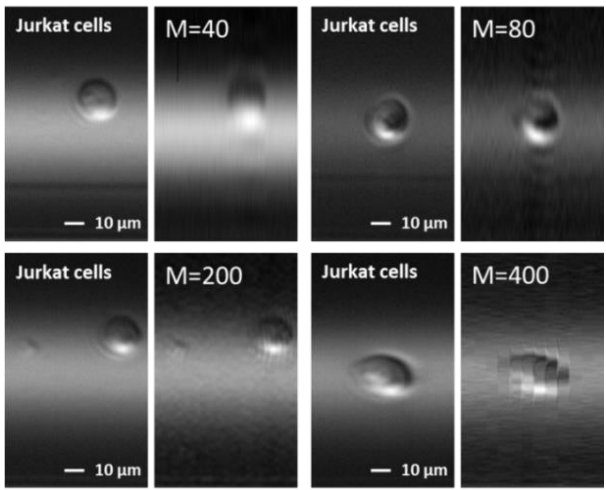

**Fig. 8.** Effect of the number of measurements on image reconstruction quality.

### C. Real-time image reconstruction

Compared with the time-stretch-based compressive imaging systems, the primary advantage of our scheme is its capability of performing image reconstruction in real time. To verify this property, we compare the computational time of image reconstruction using our scheme with that using CS-based imaging systems. When imaging flowing cells, the number of measurements captured in these two imaging systems is set to 80 ($M$) and the number of pixels the original 1-D image has is 800 ($N$). Then different recovery algorithms, two-step iterative shrinkage/thresholding algorithm (TWIST), gradient projection for sparse reconstruction (GPSR) for the CS-based imaging system and IDFT operation for our scheme, are used to reconstruct images. Now we analyze the computational cost of the above-mentioned algorithms. The TWIST algorithm and the GPSR algorithm are both iterative optimization approaches. According to [35-39], the computational cost of each iteration of TWIST and GPSR is $O(N\log(N))$ and $O(M \cdot N)$, respectively. Besides, accurately predicting the number of iterations required to find an approximate solution is impossible in terms of TWIST and GPSR [38, 39]. However the number of iterations is related to the scale of the sensing matrix and the specified error threshold. To achieve a threshold of $10^{-4}$ (common in image reconstruction), lots of iterations are needed. In our scheme, no iterative algorithm is required and only an IDFT operation is performed. The computational cost of such an operation is $O(M\log(M))$. Therefore, the computational complexity involved in our scheme is several orders of magnitude lower than that of the CS-based scheme. Next we experimentally verify the analysis mentioned above. On condition that the recovery error is fixed, the computational time for recovering 1-D images with different algorithms is calculated. We do 52 reconstructions and the results in Fig. 9 show that the computational time using the GPSR algorithm is around 0.3 s and the time is reduced to about 0.03 s if we use the TWIST algorithm. But in our scheme, only IDFT is applied to the captured measurements, the computational time is two orders of magnitude lower, approximately $2\times10^{-4}$ s. All these operations are performed by using MATLAB (Intel Core i3-3220 CPU @ 3.30 GHz, 4.00 GB RAM). The reconstruction will be significantly accelerated if high-speed digital signal processors are used. Therefore, our scheme is a promising solution to real-time imaging of flowing cells.

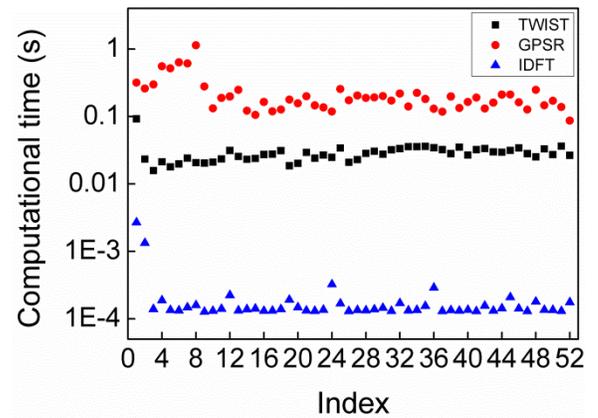

**Fig. 9.** Comparison of the reconstruction time using different recovery algorithms.

### 5. CONCLUSIONS

In summary, we demonstrate real-time high-speed single-pixel imaging by using our scheme. Phase-shifting sinusoidal structured illumination is applied for spatial spectrum acquisition of the image and inverse Fourier transform is performed on the obtained spectrum to yield high-quality images. A proof-of-concept implementation with a frame rate of 625 kHz and a compression ratio of 10% is demonstrated by imaging and reading the QR codes. To show the utility of our prototype system, an imaging flow cytometer with a throughput of 100,000 cells/s for high-throughput screening is also demonstrated. Moreover, our scheme provides an approach to achieve real-time image reconstruction that is impossible to implement with the previous CS-based imaging scheme. Our proposed imaging system not only offers the capability of observing dynamic phenomena, but also relieves pressure on data transmission, processing and storage in real-time imaging, which can be widely applied in industrial quality control and label-free biomedical imaging.

**Funding sources and acknowledgments.** This work is supported by NSFC under Contracts 61120106001, 61322113, 61271134; by the young top-notch talent program sponsored by Ministry of Organization, China; by Tsinghua University Initiative Scientific Research Program.